\documentclass[10pt,twocolumn,letterpaper]{article}

\usepackage{multirow} 
\usepackage{cvpr}              

\definecolor{cvprblue}{rgb}{0.21,0.49,0.74}
\usepackage[pagebackref,breaklinks,colorlinks,allcolors=cvprblue]{hyperref}

\def\paperID{1234} 
\def\confName{CVPR}
\def\confYear{2026}

\title{Hard to See, Hard to Label: Generative and Symbolic Acquisition for Subtle Visual Phenomena}

\author{
Renjith Prasad$^{1,2}$\thanks{Equal contribution.}\hspace{0.5em}
Rishabh Sharma$^{2}$\footnotemark[1]\hspace{0.5em}
Andrew E. Shao$^{2}$\hspace{0.5em}
Annmary Justine Koomthanam$^{2}$\\
Shreyas Kulkarni$^{2}$\hspace{0.5em}
Suparna Bhattacharya$^{2}$\hspace{0.5em}
Martin Foltin$^{2}$\hspace{0.5em}
Amit Sheth$^{1,3}$\\
David Orozco$^{4}$\hspace{0.5em}
Matthew Quinn$^{4}$\hspace{0.5em}
Brian Sammuli$^{4}$\\[0.5em]
$^{1}$University of South Carolina\\
$^{2}$AI Research Lab, HPE Labs, Hewlett Packard Enterprise\\
$^{3}$Indian AI Research Organization\\
$^{4}$General Atomics
}

\begin{document}
\maketitle

\begin{abstract}
Subtle visual anomalies such as hairline cracks, sub-millimeter voids, and low-contrast inclusions are structurally atypical yet visually ambiguous, making them both difficult to annotate and easy to overlook during active learning. Standard acquisition heuristics based on discriminative uncertainty or feature diversity often overselect dominant patterns while underexploring sparse yet important regions of the data space. This failure mode is especially severe in industrial defect inspection, where anomalies may be both low-prevalence and difficult to distinguish from surrounding structure. To resolve this, we propose \textbf{GSAL}, an active learning framework for object detection that combines a diffusion-based difficulty signal with a hierarchical semantic coverage prior. The diffusion component scores images and proposals using reconstruction discrepancy and denoising variability, prioritizing visually atypical or ambiguous examples. However, diffusion alone does not prevent acquisition from repeatedly favoring hard samples within dominant semantic modes. The semantic component therefore organizes candidate samples in a three-level concept graph and promotes coverage of underrepresented semantic regions while providing interpretable acquisition rationales. By balancing visual difficulty with semantic coverage, GSAL improves retrieval of subtle and rare targets that are often missed by uncertainty-only selection. Experiments on a proprietary thin-film defect,  Pascal VOC and  MS COCO dataset show consistent gains in label efficiency and rare-class retrieval over uncertainty-, diversity-, and hybrid-based baselines.
\end{abstract}
\raggedbottom

\section{Introduction}

\label{sec:intro}

\noindent\begin{figure}[t]
    \centering
    \includegraphics[width=0.47\textwidth]{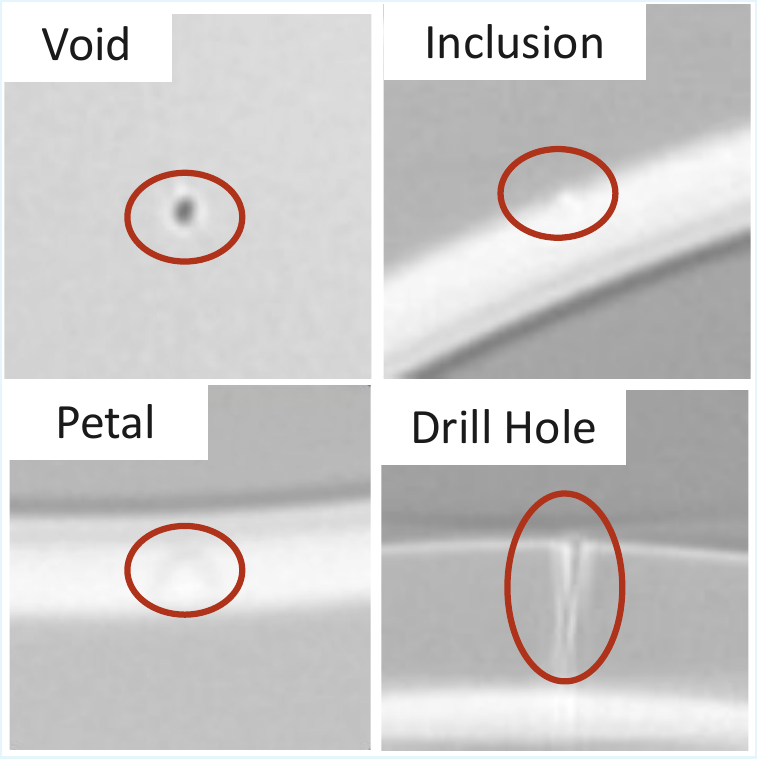}
  \caption{\textbf{Qualitative examples from the industrial thin-film defect dataset.} Representative defect types include \emph{inclusion}, \emph{drill hole}, \emph{Petal}, and \emph{void}. These defects are often small, weakly contrasted, and visually confounded with background texture, making them easy to overlook under standard active learning heuristics.}
    \label{fig:qual}
\end{figure}

 Current AL methods for object detection rely on discriminative uncertainty~\cite{lewis1994sequential, gal2017deep, beluch2018power}, feature-space diversity~\cite{sener2018active, ash2019deep}, or hybrids~\cite{yuan2021multipleinstanceactivelearning, yang2024plugplayactivelearning, hekimoglu2023activelearningobjectdetection}. All share a common assumption: that discriminative confidence, or distance in learned feature space, reliably proxies annotation value. For subtle visual anomalies this assumption fails in three compounding ways.
 \begin{figure*}[t]
    \centering
    \includegraphics[width=0.7\textwidth]{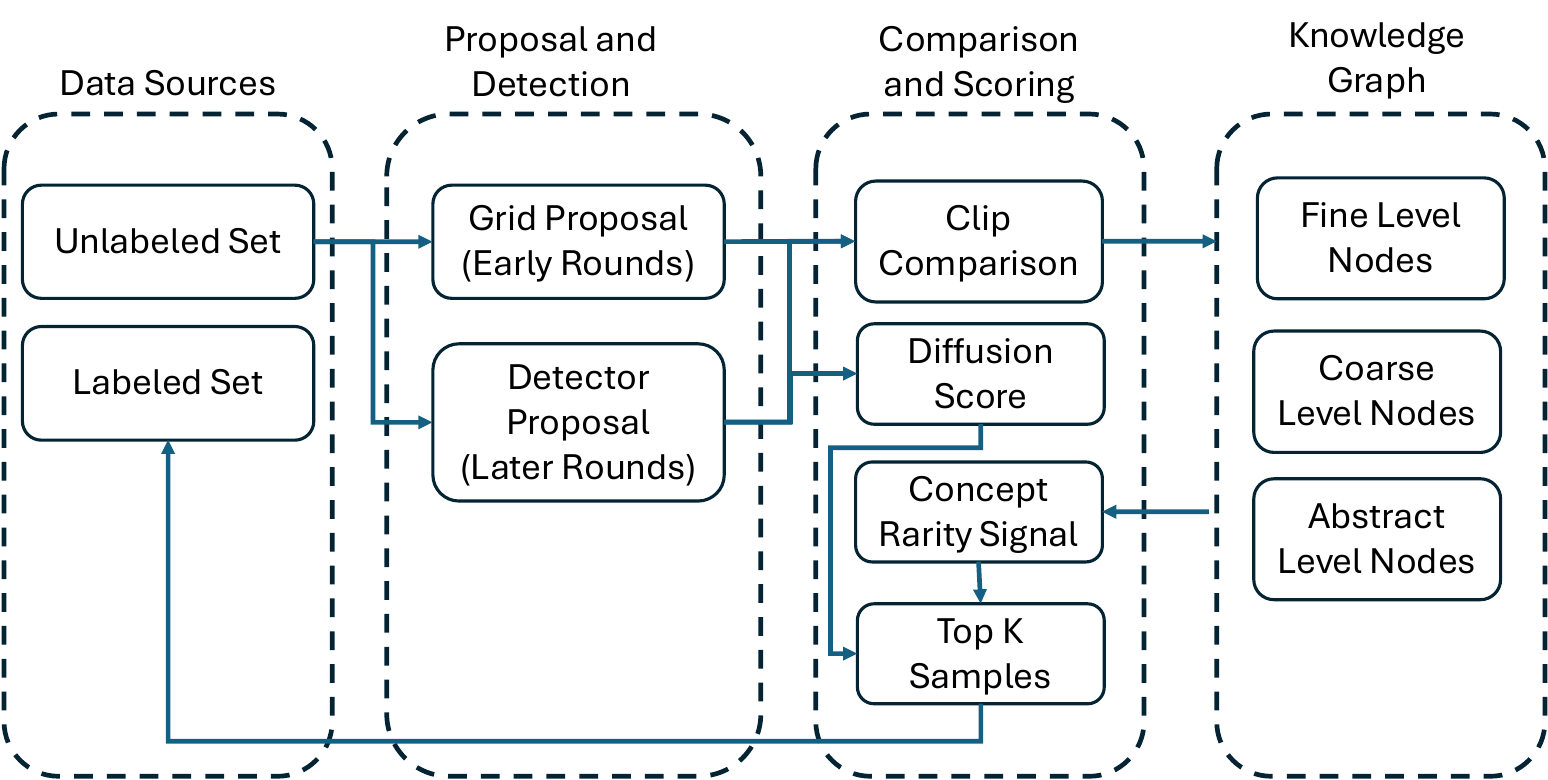}
\caption{
Overview of \textbf{GSAL}. Given an unlabeled image, the detector first generates region proposals. We then compute diffusion-based difficulty scores at two levels: on the full image to capture global scene ambiguity, and on each proposal to capture localized subtle anomalies that may be diluted at image level. In parallel, each proposal receives a rarity-aware bonus from a three-level concept graph that tracks underrepresented semantic regions across acquisition rounds. Proposal-level difficulty and coverage scores are normalized and aggregated via soft top-$k$ pooling, then fused with the image-level diffusion score to produce the final acquisition score for the image. This design allows GSAL to distinguish unusual-but-common patterns from unusual-and-rare targets, while also providing a human-readable semantic rationale for selection.
}
\label{fig:GSAL_architecture}
\end{figure*}

 Subtle visual anomalies are defined by a property that makes them more resistant to standard active learning: they are structurally atypical yet visually indistinguishable from background, occupying small spatial regions and appearing rarely in unlabeled data pools. In industrial thin-film inspection, for example, hairline drill holes and sub-millimeter voids occupy fraction of an image while closely mimicking background texture. This combination of visual ambiguity and semantic rarity creates a compounding failure for annotation-efficient learning: a detector trained on few examples confidently assigns subtle anomalies to background, assigning them low acquisition priority and leaving the labeled pool biased toward visually salient categories. Active learning is a natural tool for interrupting this dynamic, yet we argue that the dominant assumptions of existing AL strategies are violated precisely where subtle visual anomalies live. Representative defect types from the industrial thin-film defect dataset are shown in Figure~\ref{fig:qual}, illustrating the small scale, weak contrast, and morphological variability that make this setting challenging for active learning.

First, discriminative uncertainty estimates such as softmax entropy or dropout variance fail in low-label regimes; subtle anomalies that resemble background receive low entropy and low acquisition priority, because the detector has not yet learned to separate them~\cite{beluch2018power, ovadia2019trustmodelsuncertaintyevaluating, riedlinger2022gradientbasedquantificationepistemicuncertainty}. Second, diversity in latent feature space does not guarantee coverage of semantically important but sparse concepts; subtle or rare targets are underexplored when they appear close to dominant background patterns~\cite{roy2018deep, Liang_2023}. Finally, in auditable inspection workflows where missed defects carry operational and regulatory consequences, selection decisions must be traceable. An operator cannot certify the AL system without a rationale for each acquisition.

These failures point to a deeper issue: even hybrid AL strategies usually combine signals tied to detector confidence or geometric spread in feature space. That is not enough for subtle anomalies. A sample can be visually atypical yet receive low discriminative uncertainty, or maybe challenging for the generative model despite falling within the semantic mode that is already well represented in the labeled pool. Effective querying therefore requires a more targeted hybrid: one signal should capture visual atypicality, while the other should promote coverage of semantically sparse but important regions.

Diffusion models provide a natural source of generative difficulty for subtle-target acquisition.  By measuring reconstruction error and denoising variance in latent space, they assign high scores to structurally atypical, visually ambiguous regions independently of the detector's state~\cite{rombach2022high, nichol2021improveddenoisingdiffusionprobabilistic, berry2024sheddinglightlargegenerative}. But generative difficulty alone is insufficient: unusual but frequent background patterns may also score highly without corresponding to underrepresented targets worth prioritizing. A hierarchical concept graph supplies the missing semantic coverage prior, distinguishing unusual-but-common patterns from semantically sparse but important phenomena while providing a traceable rationale for selection~\cite{sheth2023neurosymbolicaiwhy, besold2017neural, garcez2019neurosymbolic}.

We propose \textbf{GSAL} (\underline{G}enerative and \underline{S}ymbolic  \underline{A}ctive \underline{L}earning), a hybrid active learning framework for object detection. GSAL applies diffusion-based scoring at both the image and proposal levels, making acquisition sensitive not only to global scene ambiguity but also to spatially localized anomalies. To distinguish unusual-but-common patterns from unusual-and-rare ones, it augments this generative signal with a three-level concept graph (\emph{fine} $\rightarrow$ \emph{coarse} $\rightarrow$ \emph{abstract}) that tracks semantic coverage across rounds and assigns a rarity-aware bonus to underrepresented regions of the pool. The resulting acquisition rule balances visual difficulty with semantic coverage while providing a human-readable rationale for selection. Our contributions are:

\begin{itemize}
    \item \textbf{Proposal-level diffusion scoring} that localizes 
    generative uncertainty to individual detector proposals via 
    generalized-mean aggregation of reconstruction error and 
    denoising variance, capturing subtle regional ambiguity that 
    image-level scoring dilutes.
     
    \item \textbf{Rarity-aware semantic coverage} through a
    three-level concept graph that steers acquisition toward rare,
    underrepresented targets instead of repeatedly selecting
    hard-but-common samples, with each selection grounded in a human-readable semantic path auditable in safety critical workflows.
 
    \item \textbf{A unified acquisition function for subtle anomaly detection} that jointly addresses visual atypicality and semantic rarity at the proposal level with a scale-normalized combining rule that is consistent across datasets without per-dataset tuning.
    \end{itemize}

On a thin-film CT defect benchmark where anomalies occupy under 
2\% of image area, GSAL achieves F1 of 0.770 at 10\% labeling 
versus 0.756 for the strongest baseline, with the largest margins 
in the 1--5\% labeling regime where subtle anomaly annotation is 
most costly. We further validate on Pascal VOC (71.1 mAP@50 at 
20\% labeling versus 63.1 for entropy and 69.5 for the strongest 
hybrid) and MS COCO (31.5 AP@[.5:.95] versus 29.6 for entropy), 
recovering all rare classes by 15\% labeling where the strongest 
baseline recovers five of six.
\section{Related Work}

Subtle visual anomalies, including industrial defects, camouflaged objects, and low-contrast medical findings, are characterized by low contrast with background, small spatial extent, and semantic rarity~\cite{shyalika2024comprehensive}. Standard detectors and foundation models~\cite{kirillov2023segment, schafer2024overcoming} underperform in this regime, making annotation cost the primary bottleneck~\cite{prasad2024assemai}.

Active learning mitigates annotation cost but fails systematically on subtle phenomena. Uncertainty-based methods~\cite{lewis1994sequential, gal2017deep, beluch2018power} are miscalibrated precisely where subtle anomalies arise: a detector with few subtle examples confidently assigns them to background~\cite{ovadia2019trustmodelsuncertaintyevaluating, riedlinger2022gradientbasedquantificationepistemicuncertainty}. Diversity-based methods~\cite{sener2018active, ash2019deep} provide no semantic coverage guarantee for anomalies geometrically indistinguishable from background in feature space. Hybrids~\cite{yuan2021multipleinstanceactivelearning, yang2024plugplayactivelearning, hekimoglu2023activelearningobjectdetection} inherit both failure modes for rare, visually ambiguous categories.

Diffusion models offer uncertainty signals grounded in structural atypicality rather than classifier confusion~\cite{berry2024sheddinglightlargegenerative, rombach2022high}, but existing diffusion-based AL operates at the image level and targets classification or domain adaptation~\cite{kushnir2020diffusionbaseddeepactivelearning, du2023diffusion, barba2025diffusionactivelearningdatadriven}, leaving subtle detection unaddressed.

Neurosymbolic methods~\cite{besold2017neural, garcez2019neurosymbolic, sheth2023neurosymbolicaiwhy} and concept bottleneck models~\cite{koh2020conceptbottleneckmodels} demonstrate that grounding neural predictions in explicit symbolic structures improves both semantic coverage and interpretability. Combined with vision-language grounding via CLIP~\cite{radford2021learning}, symbolic reasoning enables label-free concept discovery and traceable rationales, yet remains unexplored in active learning for subtle visual detection. GSAL closes this gap by coupling proposal-level diffusion uncertainty with a CLIP-grounded three-level concept graph that jointly resolves visual atypicality and semantic underrepresentation, bringing both generative and symbolic signals to bear on the subtle anomaly detection problem.
\section{Problem Statement}

Let $\mathcal{D} = \mathcal{L} \cup \mathcal{U}$ denote a pool-based active learning setting, where the labeled set is $\mathcal{L} = \{(x_i, y_i)\}_{i=1}^n$ and the unlabeled pool is $\mathcal{U} = \{u_j\}_{j=1}^m$. Within $\mathcal{U}$, we are particularly interested in an unknown subset $\mathcal{A} \subseteq \mathcal{U}$ containing subtle visual anomalies that are spatially small, low-contrast, and rare. The goal is to select a batch $\mathcal{S} \subseteq \mathcal{U}$, with $|\mathcal{S}| = k$, whose annotation leads to the largest improvement in detector performance:
\begin{equation}
\mathcal{S}^\star = \arg\max_{\mathcal{S} \subseteq \mathcal{U}, \; |\mathcal{S}| = k} \Delta \mathcal{P}(\mathcal{M}; \mathcal{S}),
\end{equation}
where $\Delta \mathcal{P}(\mathcal{M}; \mathcal{S})$ denotes the expected improvement of detector $\mathcal{M}$ after labeling and training on $\mathcal{S}$. In subtle-anomaly regimes, an effective acquisition rule should preferentially surface samples from $\mathcal{A}$, even though membership in $\mathcal{A}$ is not observed during selection.

In practice, this objective is intractable and must be approximated by an acquisition surrogate. Standard surrogates based on discriminative uncertainty~\cite{gal2017deep} or feature-space diversity~\cite{sener2018active} can be insufficient in subtle-anomaly regimes. Low-contrast targets that resemble background may receive low discriminative uncertainty despite being important to annotate~\cite{ovadia2019trustmodelsuncertaintyevaluating, riedlinger2022gradientbasedquantificationepistemicuncertainty}, while geometric spread in feature space does not guarantee coverage of semantically sparse but important categories~\cite{sener2018active, roy2018deep, Liang_2023}.

To address this, we use a hybrid acquisition score
\begin{equation}
S(u) = \hat{U}_g(u) + \eta \hat{B}(u),
\end{equation}
where $\hat{U}_g(u)$ is a diffusion-based generative difficulty score and $\hat{B}(u)$ is a semantic coverage bonus derived from a hierarchical concept graph. The first term prioritizes visually atypical or ambiguous samples, while the second promotes coverage of underrepresented semantic regions. Together, they form a practical acquisition surrogate tailored to subtle and underrepresented detection settings.
\section{Methodology}
\label{methodology}

We approximate the intractable active-learning objective in Eq.~1 with a hybrid acquisition rule designed for subtle visual anomalies. Our setting is characterized by defects that are spatially small, low-contrast, and semantically sparse, where standard discriminative uncertainty and feature-space diversity are often unreliable. To address this, our method combines two complementary signals: (i) a \emph{generative difficulty} score that surfaces visually atypical or hard-to-reconstruct samples, and (ii) a \emph{symbolic coverage} score that promotes underrepresented defect concepts. Together, these components favor samples that are both hard to see and important to annotate.

\subsection{Multi-Level Concept Graph}

To make semantic coverage explicit, we construct a three-level directed acyclic graph (DAG)
\[
\mathcal{G} = (V, E),
\]
with node set
\[
V = V_{\text{fine}} \cup V_{\text{coarse}} \cup V_{\text{abstract}},
\]
where fine-level nodes correspond to dataset labels, coarse-level nodes group related defect families, and abstract-level nodes encode shared visual attributes.

\begin{figure}[t]
    \centering
    \includegraphics[width=0.47\textwidth]{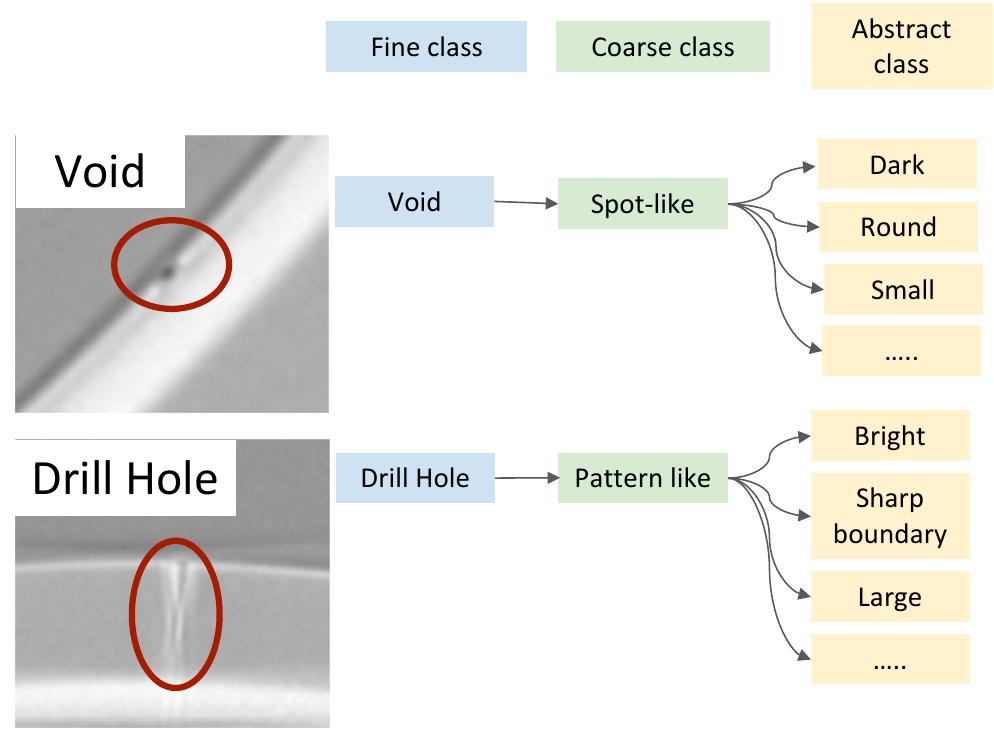}
    \caption{\textbf{Representative concept graph for the industrial thin-film defect dataset.} The figure illustrates how defect types such as \emph{void} and \emph{drill hole} are organized across three semantic levels in GSAL: fine labels, coarse morphological groups, and non-exclusive abstract attributes. This hierarchy makes semantic underrepresentation explicit at multiple granularities and supports interpretable acquisition rationales.}
    \label{fig:concept_graph}
\end{figure}

This three-level structure is intended to separate three different notions of coverage. Fine nodes preserve fidelity to the annotated label space; coarse nodes prevent repeated acquisition from the same broad defect family; and abstract nodes capture reusable visual traits such as polarity, shape, or boundary structure. For the proprietary inspection dataset, for example, fine classes include \emph{void}, \emph{inclusion}, \emph{petal}, and \emph{drill hole}; coarse groups capture broader morphological families; and abstract attributes describe properties such as \emph{dark}, \emph{bright}, \emph{round}, \emph{u-shaped}, or \emph{sharp-boundary}. This makes semantic gaps visible at multiple granularities rather than leaving them implicit in latent feature space. A representative portion of this hierarchy for the industrial thin-film defect dataset is shown in Figure~\ref{fig:concept_graph}, illustrating how fine defect labels are connected to broader morphological groups and non-exclusive abstract attributes.

Each unlabeled sample $u \in \mathcal{U}$ is connected to the graph using CLIP~\cite{radford2021learning}. Let $\phi(u)$ denote the CLIP image embedding and $\psi(c)$ the text embedding of concept $c$. We assign the sample to a fine-level concept by
\begin{equation}
v_f(u) = \arg\max_{c \in V_{\text{fine}}} \langle \phi(u), \psi(c) \rangle.
\end{equation}
Its associated coarse and abstract nodes are then retrieved from the graph. In this way, every acquisition can be explained through a semantic path from label to family to visual attributes, which is useful in auditable inspection workflows.

To track semantic balance, we maintain a coverage count $n(v)$ for each graph node $v \in V$, indicating how often that concept has been selected so far.

\subsection{Rarity-Aware Symbolic Bonus}

We use the concept graph to reward underrepresented semantic regions during acquisition. Let $n(v)$ denote the current coverage count of node $v$, and let $\tau_f$, $\tau_c$, and $\tau_a$ denote rarity thresholds at the fine, coarse, and abstract levels. For a proposal $u$, we define
\[
\begin{aligned}
I_f(u) &= \mathbb{I}[\,n(v_f(u)) < \tau_f\,], \\
I_c(u) &= \mathbb{I}[\,n(v_c(u)) < \tau_c\,], \\
I_a(u) &= \mathbb{I}[\,n(v_a) < \tau_a\,].
\end{aligned}
\]

Here, $v_f(u)$ and $v_c(u)$ are the fine and coarse concepts assigned to $u$, and $v_a \in \mathcal{A}(u)$ indexes its abstract attributes. The resulting symbolic bonus is
\begin{equation}
B(p) = I_f(p) + I_c(p) + \lambda \sum_{v_a \in \mathcal{A}(u)} I_a(u),
\end{equation}
where $\lambda$ controls the contribution of abstract-level rarity.

This bonus favors samples that improve semantic coverage at multiple levels: rare defect identities, underexplored defect families, and infrequent visual attributes.

\subsection{Diffusion-Based Generative Difficulty}
\label{sec:diffusion_uncertainty}

Coverage alone does not guarantee that selected samples are visually informative. To capture the ambiguity of subtle anomalies, we introduce a diffusion-based generative difficulty score using a latent diffusion model~\cite{rombach2022high}. The intuition is that subtle defects are often difficult to reconstruct faithfully, even when a detector assigns them low discriminative uncertainty.

Let $u$ denote an unlabeled image, and let $z(u)\in\mathbb{R}^d$ be its latent representation under the diffusion autoencoder. Starting from noisy versions of $z(u)$, the diffusion model produces $T$ denoised reconstructions
\[
\hat z^{(1)}(u),\hat z^{(2)}(u),\dots,\hat z^{(T)}(u),
\]
where $\hat z^{(t)}(u)$ is the reconstructed latent at denoising step $t$. We summarize these reconstructions by their empirical mean
\[
\bar z(u)=\frac{1}{T}\sum_{t=1}^{T}\hat z^{(t)}(u).
\]

From these trajectories, we extract two signals. First, the reconstruction error
\begin{equation}
R(u)=\|z(u)-\bar z(u)\|_2^2
\end{equation}
measures how well the latent representation can be recovered on average. Second, the denoising variance
\begin{equation}
V(u)=\frac{1}{T}\sum_{t=1}^{T}\|\hat z^{(t)}(u)-\bar z(u)\|_2^2
\end{equation}
measures instability across denoising trajectories. We then define the image-level generative difficulty as
\begin{equation}
U_{\mathrm{img}}(u)=\alpha R(u)+\beta V(u),
\end{equation}
where $\alpha,\beta\ge 0$ control the relative contributions of reconstruction error and trajectory instability.

To make this score detection-aware, we also evaluate it on detector proposals. Let $P(u)=\{p_1,\dots,p_N\}$ be the candidate regions predicted by the current detector on image $u$. After non-maximum suppression, we retain the top-$M$ proposals,
\[
P_M(u)\subseteq P(u), \qquad |P_M(u)|=M.
\]
For each proposal crop $p\in P_M(u)$, we compute the same diffusion-based score,
\begin{equation}
U_{\mathrm{prop}}(p)=\alpha R(p)+\beta V(p),
\end{equation}
where $R(p)$ and $V(p)$ are defined  on the proposal crop.

Since acquisition is performed at the image level, proposal scores must be aggregated into a single image score. Let $\tilde U_{\mathrm{prop}}(p)$ denote the normalized proposal-level generative score and let $\tilde B_{\mathrm{prop}}(p)$ denote the normalized symbolic bonus associated with proposal $p$. We use generalized-mean pooling:
\begin{equation}
S_{\mathrm{prop}}(u)
=
\left(
\frac{1}{|P_M(u)|}
\sum_{p\in P_M(u)}
\bigl(\tilde U_{\mathrm{prop}}(p)+\lambda \tilde B_{\mathrm{prop}}(p)\bigr)^{\rho}
\right)^{1/\rho}
\end{equation}
where $\rho \ge 1$ controls the pooling behavior and $\lambda$ weights the contribution of proposal-level symbolic rarity. When $\rho=1$, this reduces to mean pooling; larger $\rho$ places more emphasis on high-scoring proposals.

We then combine local proposal evidence with global image-level difficulty:
\begin{equation}
U_g(u)=(1-\gamma)S_{\mathrm{prop}}(u)+\gamma \tilde U_{\mathrm{img}}(u),
\end{equation}
where $\tilde U_{\mathrm{img}}(u)$ is the normalized image-level generative difficulty and $\gamma\in[0,1]$ balances proposal-level and image-level information.

In the final acquisition rule, the normalized generative score is instantiated as
\[
\hat U_g(u)=S(u).
\]
Thus, the diffusion model contributes a detection-aware estimate of how visually atypical, unstable, or difficult to reconstruct an image and its candidate regions are, which is particularly useful in subtle-anomaly regimes.

\subsection{Final Acquisition Score}

Before fusion, all score components are normalized within the current acquisition round using z-score normalization. For any score term $x$, we compute
\begin{equation}
\tilde{x} = \frac{x-\mu_x}{\sigma_x+\epsilon},
\end{equation}
where $\mu_x$ and $\sigma_x$ are the mean and standard deviation of that term over the current unlabeled pool, and $\epsilon$ is a small constant for numerical stability.

We then combine the generative and semantic terms into a single acquisition score, $S(u)=\tilde{U}(u)+\eta \tilde{B}(u)$, where $\tilde{U}(u)$ denotes the normalized diffusion-based score, $\tilde{B}(u)$ the normalized semantic rarity bonus, and $\eta$ their relative trade-off. The full acquisition procedure for one active-learning round is summarized in \ref{fig:gsal_algorithm}.

\begin{figure}[t]
\centering
\fbox{%
\begin{minipage}{0.97\linewidth}
\small
\textbf{GSAL acquisition for one round of subtle-anomaly discovery}\vspace{4pt}
\begin{enumerate}[leftmargin=*]
    \item For each unlabeled image $u\in\mathcal{U}$, compute the image-level generative difficulty $U_{\mathrm{img}}(u)$.
    \item Generate detector proposals $P_M(u)$; for each proposal $p$, assign semantic concepts through $\mathcal{G}$, compute the symbolic bonus $B(p)$, and compute proposal-level generative difficulty $U_{\mathrm{prop}}(p)$.
    \item Normalize $U_{\mathrm{img}}(u)$, $U_{\mathrm{prop}}(p)$, and $B(p)$ within the current acquisition round using z-score normalization so that all terms lie on comparable scales.
    \item Aggregate proposal scores into $S_{\mathrm{prop}}(u)$, then combine them with image-level difficulty to obtain the final score $S(u)$.
    \item Select the top-$k$ images by $S(u)$, annotate them, and update graph counts for the next round.
\end{enumerate}
\end{minipage}%
}
\caption{\textbf{GSAL acquisition procedure for one active-learning round.}}
\label{fig:gsal_algorithm}
\end{figure}
\section{Experimental Setup}

\subsection{Datasets and Protocols}

Our primary evaluation is on a proprietary thin-film CT defect dataset of approximately 15,000 grayscale images, split 80/10/10 into train/val/test. The dataset contains four defect types, inclusions, voids, petals, and drill holes that are small in scale, low-contrast with surrounding material, and structurally similar to background noise, making annotation particularly costly and label efficiency critical. Ground-truth labels are binary (defect present vs.\ no defect); for concept graph construction, we use CLIP-predicted defect-type labels as fine-level nodes with materials-domain prompts (\emph{void}, \emph{petal}, \emph{inclusion}, \emph{drill hole}), coarse nodes grouping defects by structural origin (surface vs.\ subsurface), and abstract nodes capturing appearance descriptors (linear, spherical, foreign). CLIP similarity scores on this out-of-distribution domain are validated against held-out detector loss to verify monotonic alignment. Due to high annotation cost, we initialize with 1\% labeled data and increase by 1\% per round up to 10\%. Results are reported as means over three independent runs to account for stochastic variation in the low-budget regime.

We further validate on Pascal VOC 2012~\cite{everingham2010pascal} and MS COCO 2017~\cite{lin2014microsoft} to assess generalization to standard long-tailed detection benchmarks. For VOC we use \texttt{trainval2007}+\texttt{trainval2012} as the unlabeled pool and evaluate on \texttt{test2007}; for COCO, \texttt{train2017} as the pool and \texttt{val2017} for evaluation. Following standard protocols~\cite{gal2017deep, sener2018active}, we initialize with 5\% labeled data and add 2.5\% per round. VOC and COCO results are single-run, as variance across these larger pools is negligible.

\subsection{Model and Training Setup}

We adopt Faster R-CNN~\cite{ren2015faster} with a ResNet-50 backbone~\cite{he2016deep}, retrained from scratch at each AL round to avoid bias accumulation, following standard AL detection protocols~\cite{yuan2021multipleinstanceactivelearning}. Training uses SGD (lr=0.02, momentum=0.9, weight decay=$10^{-4}$, batch size=16) with images resized to 600 pixels on the short side. For semantic alignment we employ CLIP~\cite{radford2021learning} and for generative uncertainty Stable Diffusion v2.1-base~\cite{rombach2022high}. All experiments run in PyTorch on four NVIDIA A100 GPUs.

\subsection{Baselines and Metrics}

We compare against random selection; uncertainty-based methods (entropy~\cite{wu2022entropybasedactivelearningobject}, margin, MC-dropout~\cite{gal2017deep}); diversity-based core-set selection~\cite{sener2018active}; and hybrid methods (BADGE~\cite{ash2019deep}, MIAOD~\cite{yuan2021multipleinstanceactivelearning}, PPAL~\cite{yang2024plugplayactivelearning}, NORIS~\cite{hekimoglu2023activelearningobjectdetection}). Two ablation controls isolate our components: diffusion-only (no symbolic coverage) and uniform-by-label (per-class quotas, no diffusion). All baselines use a unified codebase with identical detectors, augmentations, and budgets. Performance is reported as F1 and mAP@50 for the defect dataset, mAP@50 for VOC, and mAP@[.5:.95] for COCO.

\begin{table*}[t]
\centering
\resizebox{\textwidth}{!}{
\begin{tabular}{l|cccccccccc|ccccccc|ccccccc}
\toprule
 & \multicolumn{10}{c|}{\textbf{Industrial Thin-Film Defect Dataset (F1)}} 
 & \multicolumn{7}{c|}{\textbf{Pascal VOC (mAP@50)}} 
 & \multicolumn{7}{c}{\textbf{MS COCO (AP@[0.5:0.95])}} \\
\cmidrule(lr){2-11} \cmidrule(lr){12-18} \cmidrule(lr){19-25}
 & \multicolumn{10}{c|}{\textbf{Training data labeled (\%)}} 
 & \multicolumn{7}{c|}{\textbf{Training data labeled (\%)}} 
 & \multicolumn{7}{c}{\textbf{Training data labeled (\%)}} \\
\textbf{Method} 
& 1 & 2 & 3 & 4 & 5 & 6 & 7 & 8 & 9 & 10
& 5 & 7.5 & 10 & 12.5 & 15 & 17.5 & 20
& 5 & 7.5 & 10 & 12.5 & 15 & 17.5 & 20 \\
\midrule
Entropy \cite{wu2022entropybasedactivelearningobject} 
& 0.680 & 0.689 & 0.697 & 0.706 & 0.718 & 0.726 & 0.724 & 0.738 & 0.736 & 0.748
& 45.0 & 47.6 & 51.8 & 55.9 & 59.7 & 63.8 & 63.1
& 22.0 & 23.8 & 25.9 & 27.7 & 29.1 & 29.8 & 29.6 \\

Core-set \cite{sener2018active} 
& 0.684 & 0.694 & 0.703 & 0.713 & 0.723 & 0.732 & 0.730 & 0.742 & 0.741 & 0.749
& 46.0 & 48.9 & 53.7 & 57.5 & 61.4 & 65.5 & 65.0
& 23.0 & 24.8 & 26.8 & 28.7 & 29.9 & 30.8 & 30.6 \\

BADGE \cite{ash2019deep} 
& 0.686 & 0.695 & 0.706 & 0.716 & 0.725 & 0.736 & 0.734 & 0.745 & 0.744 & 0.751
& 47.0 & 49.8 & 54.8 & 58.7 & 62.5 & 66.7 & 66.2
& 23.5 & 25.3 & 27.2 & 29.1 & 30.2 & 31.1 & 30.9 \\

MIAOD \cite{yuan2021multipleinstanceactivelearning} 
& 0.687 & 0.697 & 0.709 & 0.718 & 0.728 & 0.739 & 0.737 & 0.747 & 0.749 & 0.753
& 47.5 & 50.7 & 55.7 & 59.8 & 63.6 & 67.8 & 67.1
& 24.0 & 25.9 & 27.9 & 29.5 & 30.5 & 31.6 & 31.4 \\

PPAL \cite{yang2024plugplayactivelearning} 
& 0.688 & 0.699 & 0.711 & 0.720 & 0.730 & 0.741 & 0.740 & 0.748 & 0.750 & 0.755
& 48.0 & 51.8 & 56.9 & 60.8 & 64.8 & 68.9 & 68.4
& 24.2 & 26.0 & 28.1 & 29.8 & 30.8 & 31.9 & 31.7 \\

NORIS \cite{hekimoglu2023activelearningobjectdetection} 
& 0.689 & 0.701 & 0.712 & 0.721 & 0.732 & 0.743 & 0.741 & 0.749 & 0.751 & 0.756
& 48.5 & 52.7 & 57.8 & 61.7 & 65.9 & 69.8 & 69.2
& 24.5 & 26.3 & 28.4 & 30.0 & 31.0 & 32.0 & 31.8 \\

\textbf{GSAL (Ours)} 
& \textbf{0.691} & \textbf{0.703} & \textbf{0.715} & \textbf{0.724} & \textbf{0.736} & \textbf{0.748} & \textbf{0.746} & \textbf{0.756} & \textbf{0.759} & \textbf{0.770}
& \textbf{50.0} & \textbf{54.1} & \textbf{59.3} & \textbf{63.0} & \textbf{67.6} & \textbf{71.8} & \textbf{71.1}
& \textbf{25.0} & \textbf{26.9} & \textbf{28.9} & \textbf{30.3} & \textbf{31.3} & \textbf{32.3} & \textbf{32.1} \\
\bottomrule
\end{tabular}
}
\caption{Performance comparison of active learning methods across three datasets. The industrial thin-film defect dataset is the primary benchmark for subtle visual anomalies, along with Pascal VOC and MS COCO. \textbf{Bold} indicates the best result at each labeling budget.}
\label{tab:al_results}
\end{table*}
\section{Results}

We evaluate GSAL against uncertainty-, diversity-, and hybrid-based baselines on the thin-film CT defect dataset, Pascal VOC 2012~\cite{everingham2010pascal}, and MS COCO~\cite{lin2014microsoft}. Results directly address the three failure modes from Section~\ref{sec:intro}.

\subsection{Performance vs.\ Labeling Budget}

Table~\ref{tab:al_results} shows that GSAL consistently surpasses all baselines across datasets and budgets. On the thin-film CT defect benchmark, GSAL achieves an F1 score of \textbf{0.770} at 10\% labeling versus 0.756 for NORIS and 0.748 for entropy, with the largest margins in the 1--5\% labeling regime where entropy miscalibration is most severe. On Pascal VOC, GSAL achieves \textbf{71.1} mAP@50 at 20\% labeling versus 63.1 for entropy and 69.2 for NORIS. On MS COCO, GSAL reaches \textbf{32.1} AP@[.5:.95] versus 29.6 for entropy.

Table~\ref{tab:cross_dataset} further suggests that these gains hold at a fixed labeling budget across all three settings consistently. GSAL achieves the strongest result on every dataset, with the largest margin on the industrial defect benchmark where subtle anomalies make discriminative selection most unreliable. This cross-setting consistency confirms that the combination of generative uncertainty and symbolic coverage is most effective precisely where subtlety and semantic rarity are simultaneously highest.

\subsection{Industrial Ablations and Analysis}

\paragraph{Effect of concept-graph hierarchy.}
Table~\ref{tab:conceptgraph_levels_main} shows that the full three-level graph outperforms all partial variants on the defect benchmark. Fine-only selection cannot prevent oversampling within a coarse group: acquiring many cracks does not signal that surface defects are already covered. Coarse-only loses fine-grained variety within groups; abstract-only captures appearance diversity without anchoring to specific defect types. Crucially, subtle anomalies are underrepresented at all three levels, which is why single-level coverage consistently falls short. Only full hierarchy provides complementary signals across all levels, reliably surfacing rare defects that partial variants miss.

\paragraph{Cross-setting generalization.}
Beyond single-dataset results, Table~\ref{tab:cross_dataset} reports a cross-data evaluation in which the model is trained with data from multiple datasets and then evaluated on each target benchmark. Here, ``cross-data'' denotes a multi-source training protocol designed to test whether the active learning strategy remains effective under heterogeneous training distributions GSAL achieves the strongest result on all three datasets, including \textbf{0.77} F1 on the industrial benchmark, \textbf{74.5} mAP@50 on Pascal VOC, and \textbf{33.1} AP@[0.5:0.95] on COCO. These consistent gains across industrial and standard detection settings suggest that the proposed acquisition rule remains effective beyond any single benchmark, with its clearest benefit appearing in the subtle-anomaly regime that motivates this work.

\begin{table}[t]
\setlength{\tabcolsep}{3pt}
\centering
\begin{tabular}{lccc}
\hline
Method & Proprietary & Pascal VOC & COCO \\
& (F1) & (mAP@50) & (AP@[0.5:0.95]) \\
\hline
Random    & 0.74 & 67.0 & 31.8 \\
Entropy   & 0.75 & 69.0 & 32.0 \\
Core-set  & 0.75 & 70.0 & 32.3 \\
CLIP-only & 0.74 & 69.5 & 32.1 \\
GSAL      & \textbf{0.77} & \textbf{74.5} & \textbf{33.1} \\
\hline
\end{tabular}
\caption{Cross-dataset generalization. Results on the proprietary thin-film defect dataset, Pascal VOC, and MS COCO. The proprietary dataset is evaluated at \textbf{10\%} labeling budget, whereas Pascal VOC and COCO are evaluated at \textbf{20\%}. GSAL achieves the best performance across all three targets.}
\label{tab:cross_dataset}
\end{table}

\paragraph{Computational Efficiency.}
Entropy sampling requires only a single detector forward pass and is therefore the cheapest baseline (normalized runtime $=1.0$), but it achieves only 0.748 F1 at 10\% labeling on the industrial thin-film defect dataset as shown in \ref{tab:al_results}. Core-set and hybrid strategies introduce modest additional overhead due to feature clustering and gradient-based computations ($\sim 1.1\times$), reaching roughly 0.750--0.757 F1. GSAL, despite incorporating diffusion-based scoring, adds only a 15\% runtime overhead compared to entropy sampling thanks to proposal prefiltering and lightweight screening, which reduce the number of crops passed through the diffusion model. This modest cost yields a clear performance gain: GSAL achieves \textbf{0.770} F1, improving over entropy by +0.020 and over the strongest baseline (NORIS) by +0.013 at the same labeling budget. All experiments run in PyTorch on four NVIDIA A100 GPUs.

\paragraph{Sensitivity analysis.}
To identify the optimal operating point for GSAL, we conducted a systematic stability study over all acquisition parameters on the industrial thin-film defect dataset, varying each parameter independently while holding the others fixed. Across all tested ranges, GSAL shows stable behavior: no single parameter produces more than $\pm 0.3$ F1 variation, confirming that the fixed values reported throughout this paper represent a robust and generalizable operating point rather than a dataset-specific optimum. Diffusion weights $(\alpha,\beta)$ were set so that reconstruction error $R$ and denoising variance $V$ contribute comparably after z-score normalization, remaining stable for $\alpha/\beta \in [0.5,2.0]$. The uncertainty--coverage trade-off $\eta=0.5$ remained stable for $\eta \in [0.3, 0.7]$, and rarity weight $\lambda=1.0$ varied by under 0.2 percentage points in F1 across $[0.5, 2.0]$. Generalized-mean pooling was robust across $\rho \in \{2,4\}$, prefilter size $M \in \{16,32,64\}$ balanced accuracy and runtime, and $T=4$ stochastic reconstructions produced stable uncertainty rankings. Dynamic rarity thresholds $(\tau_f, \tau_c, \tau_a)$ were recomputed each round from top-20\% coverage percentiles, adapting automatically as the labeled set grew.

\subsection{Semantic Coverage and Rare-Category Retrieval}

A central challenge in the industrial thin-film defect setting is that subtle defects are not only visually weak, but also sparsely represented across acquisition rounds. If acquisition repeatedly favors dominant visual modes, the detector may never receive enough examples of uncommon but safety-critical defects. GSAL addresses this by explicitly coupling generative difficulty with semantic coverage.

Evidence for this mechanism is clearest on standard benchmarks, where rare-category structure can be measured directly. Representative examples of the fine $\rightarrow$ coarse $\rightarrow$ abstract mappings used by the concept graph are shown in Table~\ref{tab:concept_examples}, illustrating how rare categories can be grouped by broader semantic families and shared attributes rather than treated as isolated labels. On Pascal VOC, GSAL progressively broadens coverage beyond dominant classes, while baselines oversample frequent categories. Table~\ref{tab:rare} shows that GSAL retrieves \textbf{4} rare classes by 10\% labeling and all \textbf{6} by 15\%, versus 2--5 for baselines. The ablation in Table~\ref{tab:ablation} isolates the cause: removing the semantic component reduces rare-class coverage from 6 to 3 classes on VOC and from 10 to 6 classes on COCO, while removing diffusion reduces overall performance without collapsing coverage, confirming that generative scoring and semantic coverage play complementary roles.

\begin{table}[t]
\centering
\resizebox{\columnwidth}{!}{%
\begin{tabular}{lll}
\hline
Fine class & Coarse & Abstract attributes \\
\hline
Dog   & Animal     & Furry, Outdoor (VOC context) \\
Sheep & Animal     & Furry, Outdoor (pasture) \\
Sofa  & Furniture  & Rigid, Indoor (majority use) \\
Car   & Vehicle    & Wheeled, Outdoor (road) \\
TV    & Electronic & Rigid, Indoor (functional) \\
\hline
\end{tabular}%
}
\caption{Examples of fine $\rightarrow$ coarse $\rightarrow$ abstract mappings in the concept graph. Abstract attributes capture appearance, context, or functionality, rather than exclusivity.}
\label{tab:concept_examples}
\end{table}

\begin{table}[t]
\centering
\resizebox{\columnwidth}{!}{%
\begin{tabular}{lccccc}
\toprule
 & \textbf{Random} & \textbf{Entropy} & \textbf{Core-set} & \textbf{CLIP-only} & \textbf{GSAL (Ours)} \\
\midrule
\textbf{Rare @10\%} & 1 & 2 & 2 & 3 & \textbf{4} \\
\textbf{Rare @15\%} & 3 & 3 & 4 & 5 & \textbf{6} \\
\bottomrule
\end{tabular}%
}
\vspace{3pt}
\caption{Rare class retrieval on Pascal VOC. Each cell shows the number of rare classes retrieved at 10\% and 15\% labeling. GSAL achieves full coverage of all six rare classes by 15\%.}
\label{tab:rare}
\end{table}

We treat these benchmark results as supporting evidence for the same mechanism in the industrial setting rather than direct proof of transfer. On the thin-film defect dataset, ground-truth defect type labels are unavailable and CLIP alignment is less reliable on grayscale CT, making direct coverage measurement infeasible. The F1 gains in the 1--5\% labeling regime are consistent with the concept graph surfacing rare defect types earlier than baselines, but rigorous validation needs fine-grained defect-type annotations.

\begin{table}[t]
\centering
\small
\begin{tabular}{lccc}
\toprule
\textbf{Configuration} & \textbf{2\%} & \textbf{5\%} & \textbf{10\%} \\
\midrule
Fine-only (F)          & 0.692 & 0.721 & 0.748 \\
Coarse-only (C)        & 0.694 & 0.725 & 0.752 \\
Abstract-only (A)      & 0.693 & 0.724 & 0.751 \\
Fine + Coarse (F+C)    & 0.697 & 0.729 & 0.756 \\
\textbf{Full (F+C+A)}  & \textbf{0.702} & \textbf{0.737} & \textbf{0.770} \\
\bottomrule
\end{tabular}
\caption{Effect of concept-graph hierarchy on the industrial thin-film defect dataset (F1). The full three-level graph consistently outperforms partial variants, showing that fine, coarse, and abstract semantic levels provide complementary coverage signals in subtle-anomaly regimes.}
\label{tab:conceptgraph_levels_main}
\end{table}

\begin{table}[t]
\centering
\resizebox{\columnwidth}{!}{%
\begin{tabular}{lcccc}
\hline
\multirow{2}{*}{Method} & \multicolumn{2}{c}{Pascal VOC (20\%)} & \multicolumn{2}{c}{MS COCO (20\%)} \\
\cline{2-5}
 & mAP@50 & Rare Cov. (6) & AP@[0.5:0.95] & Rare Cov. (10) \\
\hline
GSAL (full)   & \textbf{77.0} & \textbf{6} & \textbf{33.5} & \textbf{10} \\
w/o Diffusion & 70.2 & 5 & 31.5 & 8 \\
w/o Symbolic  & 68.7 & 3 & 30.7 & 6 \\
w/o Rarity    & 67.5 & 4 & 30.2 & 7 \\
\hline
\end{tabular}%
}
\caption{Ablation on Pascal VOC 2012 and MS COCO 2017 at 20\% labeling. Removing diffusion, symbolic reasoning, or rarity bonuses reduces performance and rare-class coverage.}
\label{tab:ablation}
\end{table}
\section{Future Work}

GSAL points to two concrete next steps. First, we have explored using CLIP embeddings and LLM-based semantic clustering to automate the construction of the three-level concept graph. Initial investigations show that this greatly reduces the manual effort required to apply GSAL to new datasets by using a fully data-driven ontology that can adapt to novel industrial domains. The symbolic component is thus more scalable and better suited to settings where defect taxonomies are incomplete or evolving. Second, we aim to validate the proposed mechanism more directly across additional industrial subtle-anomaly benchmarks, especially through rare-category and component ablations. This validation is currently limited by binary labels in our defect dataset, but richer defect-type annotations would enable a direct test of whether the semantic-coverage and generative-difficulty effects observed on Pascal VOC and MS COCO transfer to other defect settings.
\section{Conclusion}

We presented GSAL, a framework that combines proposal-level diffusion uncertainty with rarity-aware symbolic coverage to address structural failures of existing active learning strategies in subtle visual detection. Discriminative signals assign low acquisition priority to anomalies that are visually indistinguishable from background, while diversity sampling provides no guarantee of surfacing semantically rare phenomena. On a thin-film CT defect benchmark where anomalies occupy under 2\% of image area, GSAL achieves consistent F1 gains over entropy and hybrid baselines across all labeling budgets, with the largest margins in the 1--5\% regime where annotation cost and visual ambiguity are both high. Diffusion alone conflates unusual-but-common with unusual-and-rare, while symbolic coverage alone provides no measure of visual difficulty. Validation on Pascal VOC and MS COCO confirms that the coverage mechanism transfers to standard long-tailed benchmarks, and we are actively extending GSAL to industrial settings with richer defect-type annotations to validate the full coverage claims that binary labels currently preclude.

{
    \small
    \bibliographystyle{ieeenat_fullname}
    \bibliography{main}
}

\end{document}